\documentclass{article} 
\usepackage{iclr2023_conference,times}
\usepackage{bbm}

\usepackage{amsmath,amsfonts,bm}

\def\eqref#1{equation~\ref{#1}}

\def\1{\bm{1}}

\DeclareMathAlphabet{\mathsfit}{\encodingdefault}{\sfdefault}{m}{sl}
\SetMathAlphabet{\mathsfit}{bold}{\encodingdefault}{\sfdefault}{bx}{n}

\usepackage{hyperref}
\usepackage{url}

\newcommand{\teacher}[0]{\textsc{teacher}}
\newcommand{\student}[0]{\textsc{student}}
\newcommand{\prompt}[0]{\textsc{prompt}}

\definecolor{mypurple}{RGB}{111,61,121}

\usepackage{graphicx}

\title{Learning by Distilling Context}

\author{Charlie Snell, Dan Klein, Ruiqi Zhong \\
University of California, Berkeley, EECS Department\\
\texttt{\{csnell22, klein, ruiqi-zhong\}@berkeley.edu} \\
}

\usepackage{wrapfig}

\iclrfinalcopy 
\begin{document}

\maketitle

\begin{abstract}
Language models significantly benefit from context tokens, such as prompts or scratchpads.
They perform better when prompted with informative instructions, and they acquire new reasoning capabilities by generating a scratch-pad before predicting the final answers.
However, they do not \textit{internalize} these performance gains, which disappear when the context tokens are gone.
Our work proposes to apply context distillation so that a language model can improve itself by internalizing these gains. 
Concretely, given a synthetic unlabeled input for the target task, we condition the model on ``[instructions] + [task-input]'' to predict ``[scratch-pad] + [final answer]''; 
then we fine-tune the same model to predict its own ``[final answer]'' conditioned on the ``[task-input]'', without seeing the ``[instructions]'' or using the ``[scratch-pad]''.

We show that context distillation is a general method to train language models, and it can effectively internalize 3 types of training signals.
First, it can internalize abstract task instructions and explanations, so we can iteratively update the model parameters with new instructions and overwrite old ones.
Second, it can internalize step-by-step reasoning for complex tasks (e.g., 8-digit addition), and such a newly acquired capability proves to be useful for other downstream tasks.
Finally, it can internalize concrete training examples, and it outperforms directly learning with gradient descent by 9\% on the SPIDER Text-to-SQL dataset; furthermore, combining multiple context distillation operations can internalize more training examples than what the context window size allows.
\end{abstract}

\section{Introduction}

Recent work has shown that language models significantly benefit from context tokens.
When prompted with task definitions, language models can perform zero-shot learning \citep{wei2022finetuned, sanh2022multitask}, and the performance further improves with additional in-context examples and explanations \citep{chen-etal-2022-meta, scheurer2022learning}.
They also acquire the capability to perform more complex tasks by generating step-by-step reasoning in the context window before predicting the final answer~\citep{nye2021show, wei2022chain, zhou2022least}.

However, language models cannot \textit{internalize} these performance gains, which disappear when the context tokens are gone.
Consequently, we always need to pay extra computation for running inference on context tokens; 
this is undesirable, as sometimes the task instructions and the scratch-pad can be more than 10x longer than the actual task inputs.
Furthermore, it is unclear how to leverage the context tokens when their total length exceeds the context window size.
These shortcomings are analogous to how humans are slow at performing  complex cognitive tasks \citep{wason1974dual} and can hold only a limited amount of information in the working memory \citep{baddeley1992working}. 

\begin{figure}[]
    \centering
    \includegraphics[width=0.9\textwidth]{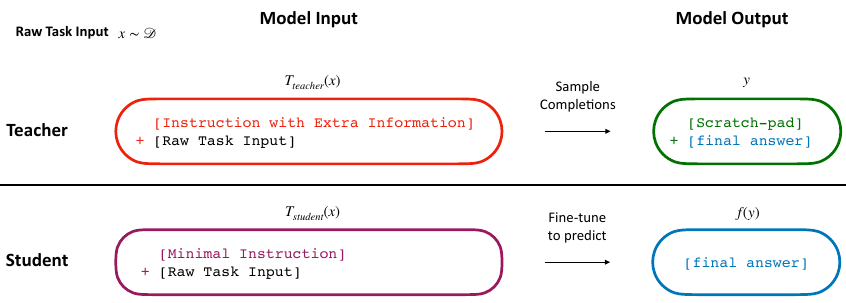}
    \caption{An overview of our context distillation framework. We sample a raw task input, form the teacher's prompt by pre-prending a detailed instruction that might contain more examples and explanations, and ask the language model to conditionally sample a scratch-pad and a final answer. Then we fine-tune the same language model to directly predict the final answer with a minimal instruction. We formalize this framework mathematically in Section \ref{sec:context-distill}. }
    \label{fig:fig1}
\end{figure}

Humans get around this by practicing. 
Consider, for example, learning to type your friends' phone numbers. 
The first few times you type it, you need to consciously recall the number using working memory and slowly decide which button to press.
After repeatedly typing the same number, it becomes a habit and you can type the number quickly without conscious reasoning.
Through repeated practice, the knowledge of your friend's phone number is ``distilled'' into your muscle memories.\footnote{See declarative learning vs. procedural learning for a friendly but more in-depth discussion. \url{https://en.wikipedia.org/wiki/Declarative_learning}}
This mechanism for distilling knowledge is critical for learning complex tasks because it allows us to incrementally build up our knowledge and skills, so that we can learn to accomplish increasingly complex tasks.

\begin{figure}[h]
    \centering
    \includegraphics[width=0.8\textwidth]{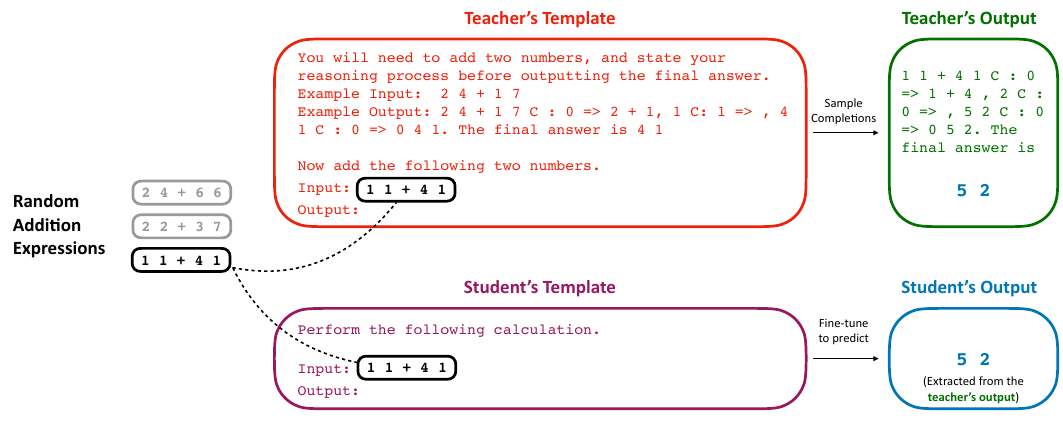}
    \caption{
    To internalize step-by-step reasoning via context distillation, we first sample a raw task input, insert it into the teacher template to form a prompt, use the language model to generate a completion with a scratch-pad in green, and extract the final answer in blue. 
    We then fine-tune the same language model to directly predict the final answer conditioned on the student prompt. 
    }
    \label{fig:step-by-step}
\end{figure}

We propose to apply a similar method, context distillation, to fine-tune language models.
For example, as shown in Figure~\ref{fig:step-by-step}, to make language models internalize the step-by-step addition capability, we first synthesize a large number of ``practice'' addition questions;
we then ask the model to follow the more informative instruction to reason step-by-step before generating the target answer; 
finally, we fine-tune the language model to directly predict the answer conditioned on a simpler student prompt.
As a result, by practicing on a lot of addition problems, the ability to add is distilled into its parameters.
We formally state our generalized context distillation framework in Section \ref{sec:context-distill}.

Section \ref{sec:experiments} shows that context distillation is a general method to train language models, and we can apply it to a wide range of settings: learning from abstract statements, learning from concrete examples, and learning from step-by-step reasoning.
Section \ref{sec:learning-from-abstract} (Figure \ref{fig:abstract}) shows that context distillation can effectively internalize task instructions from Natural-Instructions-V2 \citep{wang2022benchmarking};
it can also benefit from natural language explanations of why certain outputs are correct or incorrect;
additionally, we can teach the student to associate numerical indices with certain tasks, and then we can sequentially re-assign these task indices, overwriting the student's past associations.
Section \ref{sec:learning-from-examples} (Figure \ref{fig:concrete}) shows that context distillation can be used to internalize Text-to-SQL training examples from the SPIDER dataset \citep{yu-etal-2018-spider} into Incoder~\citep{fried2022incoder}, and it outperforms directly learning with gradient descent by 9\% for 8-shot adaptation;
additionally, we show that as we distill more training examples than can fit in the context window, we observe continual improvements in performance.
Section \ref{sec:step-by-step} (Figure \ref{fig:concrete}) shows that we can internalize step-by-step reasoning to perform 8-digit addition, and such a capability can transfer to downstream question answering tasks.

Overall, context distillation demonstrates promising potential as a general method to train language models. As discussed in Section \ref{sec:related}, we predict that future models will be better able to learn from context than today's models, and researchers will use these models to tackle increasingly complex tasks that require more extensive background knowledge and longer reasoning chains.
Therefore, we anticipate our method to be increasingly useful in the future.

\section{Context Distillation} \label{sec:context-distill}

We introduce the main components and the intuition of our context distillation framework in Section \ref{sec:intuition}, describe our algorithm for single round distillation in Section \ref{sec:algo}, explain how to distill multiple contexts sequentially or simultaneously in Section \ref{sec:multiple-updates},
and describe various implementation details to make it efficient and stable in Section \ref{sec:implementation-detail}.

\subsection{Intuition and Main Components} \label{sec:intuition}

We explain our method by contrasting it with the classical distillation methods~\citep{https://doi.org/10.48550/arxiv.1503.02531}.
These classical methods ask the teacher model with parameter $\theta_{\teacher}$ to generate a label $y$ for a given input $x$, and train the student $\theta_{\student}$ to mimic the teacher by predicting $y$ conditioned on $x$.
Typically, $\theta_{\teacher} \neq \theta_{\student}$ when the algorithm starts, and the distillation process is driven by the difference between their parameters.
In contrast, under context distillation,  $\theta_{\teacher} = \theta_{\student}$ when the training starts, and the distillation process is instead driven by the differences in the $x$ and $y$ that they see and predict.

To design such a difference that drives the distillation process, our framework requires the model developers to provide four components: a raw task input distribution $\mathcal{D}$, a teacher template $T_\teacher$, a student template $T_\student$, and an answer extractor $f$.
We introduce them below.

\paragraph{Raw Task Input Distribution $\mathcal{D}$.} 
$\mathcal{D}$ is a distribution of strings, which are typically the ``core'' inputs of the target task of interest. For example, if the target task is to classify movie review sentiment, the input distribution could be defined as random movie reviews.
More generally, there are many ways to define a raw task input distribution: we can use a rule-based method to generate random strings, sample from a pool of unlabeled data, or conditionally sample from a language model.
We explicitly distinguish raw task inputs from the whole ``input prompt'' to the language model, which is obtained after applying the templates below.

\paragraph{Teacher Template $T_\teacher$.}
$T_\teacher$ is a mapping from strings to strings, which transforms raw task inputs to the input prompts for the teacher model.
As shown in Figure \ref{fig:fig1}, the teacher template usually contains detailed instructions, explanations, and examples about the task. 

\paragraph{Student Template $T_\student$.}
$T_\student$ is a mapping from strings to strings, which transforms raw task inputs to the input prompts for the student model.
As shown in Figure \ref{fig:fig1}, this template usually still contains minimal information about the task so that the request in the prompt is not under-specified.
However, compared to the teacher prompt, it incorporates far fewer explanations and training examples of the task, and such a difference transfers this useful context information into the student parameters.

\paragraph{Answer Extractor $f$.}
$f$ is a mapping from token sequences to token sequences, which extracts the final answer (a sub-sequence of tokens) from the full teachers' generation.
As shown in Figure \ref{fig:fig1}, $f$ strips away the intermediate reasoning process, and the students need to internalize the step-by-step reasoning process to directly predict what the teacher predicts at the end.

We now describe context distillation formally using the mathematical terms we just introduced.

\subsection{Formal Description} \label{sec:algo}

Our algorithm first samples an $x$ from $\mathcal{D}$ and ask the language model to sample a completion $y$ conditioned on $T_{\teacher}(x)$; we then fine-tune the language model to predict $f(y)$ conditioned on $T_{\student}(x)$.
Throughout the distillation process, $\theta_{\teacher}$ is fixed.

Formally, let $\theta_{\student}$ and $\theta_{\teacher}$ be the parameters of a language model, and define $P_{\theta}(\cdot|\prompt)$ to be the probability distribution of the completions conditioned on the prompt. 
We optimize 

\begin{equation} \label{eq:total-loss}
    \mathcal{L}_{\mathcal{D}, T_{\student}, T_{\teacher}, f, \theta_{\teacher}}(\theta_{\student}) = \mathbb{E}_{x\sim\mathcal{D}}[\mathbb{E}_{y\sim P_{\theta_{\teacher}(\cdot|T_{\teacher})}}[\log P_{\theta_{\student}}(f(y)|T_{\student}(x))]]
\end{equation}

Notice that the definition of $\mathcal{L}$ depends on five variables $\mathcal{D}$, $T_{\student}$, $T_{\teacher}$, $f$, and $\theta_{\teacher}$. 
To keep the notation uncluttered, we will only include the necessary subscripts if the rest can be inferred from the surrounding text.

\subsection{Combining Multiple Updates} \label{sec:multiple-updates}
We now introduce simultaneous distillation and sequential distillation, two straightforward variants that combine multiple context distillation operations, allowing us to internalize ensembles or sequences of contexts, enabling a form of learning that is not possible with just a single prompt.

\paragraph{Simultaneous Distillation.} 
To simultaneously perform $K$ different context distillation operations represented by $\mathcal{D}_{1\dots K}, T_{\teacher/\student, 1\dots K}, f_{1\dots K}$, we can optimize the total loss:
\begin{equation}
    \mathcal{L}_{\textsc{total}} := \sum_{k=1}^{K} \mathcal{L}_{\mathcal{D}_{k}, T_{\student, k}, T_{\teacher, k}, f_{k}}
\end{equation}

Simultaneous distillation is especially useful when the prompts contain independent instructions and in-context training examples, but their total length exceeds the language model context window size.

\paragraph{Sequential Distillation.}

To perform $K$ context distillation operations sequentially, we can inductively define $\theta_{\student, 0}$ as the initial language model, and $\theta_{\student, k+1}$ to be the parameters after fine-tuning with the loss function
\begin{equation}
    \mathcal{L}_{k} := \mathcal{L}_{\mathcal{D}_{k}, T_{\student, k}, T_{\teacher, k}, f_{k}},
\end{equation}
with $\theta_{\student, k}$ as the initialization.
Sequential distillation is useful for incrementally updating the student model or overwriting previous updates. We can also compose these two variants arbitrarily.

\paragraph{Recursive Distillation.} Beyond the above two variants, ~\citet{https://doi.org/10.48550/arxiv.2206.11349} explores a recursive variant, which is similar to sequential distillation, except that the student model becomes the new teacher in the next iteration.
This allows us to incrementally update the model without maintaining a separate set of parameters for the student and the teacher.

\subsection{Implementation Details} 
\label{sec:implementation-detail}

A naive method to optimize Equation \ref{eq:total-loss} is to sample $y$ and directly fine-tune the student to predict the hard label of $f(y)$: such a method wastes the token logit information and results in noisy gradients. 
Instead, we minimize the token-level KL divergence between the student and the teacher.
However, this leads to another issue: the vocabulary space of language models is often on the order of 50-100k, and the full soft labels consume a lot of memory.
Therefore, we approximate the soft label distribution by an empirical distribution of 100 token samples. 
Such a technique saves us a lot of memory while still delivering satisfactory performance.

\section{Experiments} \label{sec:experiments}

We apply context distillation to three types of settings. 
In Section \ref{sec:learning-from-abstract}, we apply context distillation to internalize abstract instructions and natural language explanations; 
additionally, we show that context distillation can associate multiple task instructions with a task id, and sequential distillation can overwrite previous updates.
In Section \ref{sec:learning-from-examples}, we apply context distillation to internalize concrete training examples, and we show that it outperforms directly learning with gradient descent on the SPIDER Text-to-SQL dataset;
additionally, we show that simultaneous distillation can be used to internalize more training examples than the context window can fit.
In Section \ref{sec:step-by-step}, we apply context distillation to internalize the ability to perform 8-digit addition step-by-step;
additionally, we show that such a capability can transfer to other downstream question-answering tasks even when the scratch-pad is not present.
In all cases, the student's input length is significantly shorter than the teacher's, and we may reduce the context length by as much as 11 times, saving us a large amount of compute at inference time.

For each experiment, we report the teacher's performance and the student's performance before and after distillation.
The student's performance needs to improve after context distillation in order to support the claim that context distillation is successful.
The teacher's performance is generally an upper-bound on the student's performance, except for the case of simultaneous distillation where multiple teacher templates are applied and no individual teacher can outperform the student.

\subsection{Learning From Abstract Instructions and Explanations} \label{sec:learning-from-abstract}

\begin{figure} [t]
    \centering
    \includegraphics[width=0.8\textwidth]{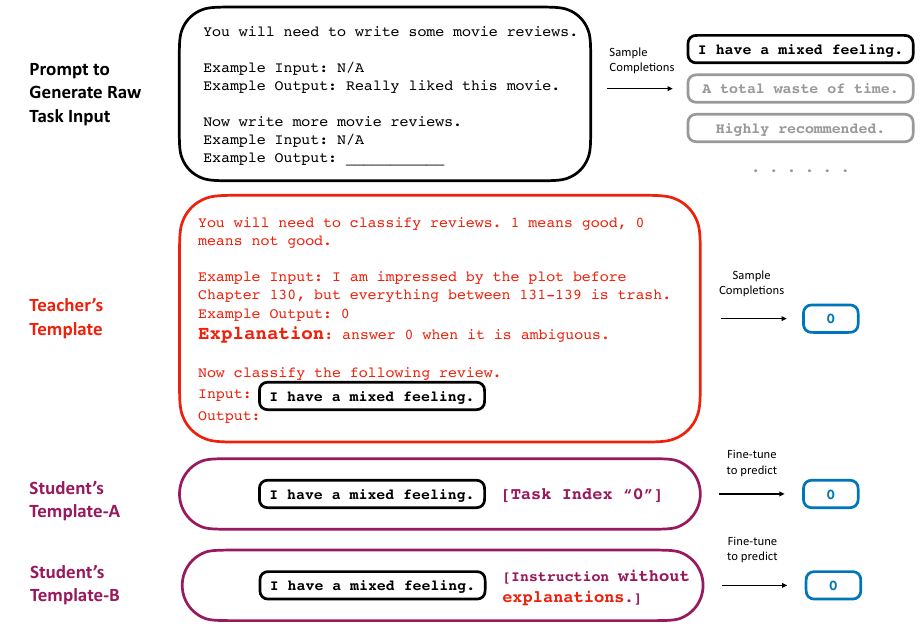}
    \caption{
    We use context distillation to internalize abstract task instructions and associate each of them with a task id. 
    For example, after distilling the context with the teacher's template and student template-A, the student should perform sentiment classification whenever it sees the index ``[0]''.
    We perform an additional ablation study by using the student template-B without the explanation to verify that context distillation can be used to learn from natural language explanations. 
    The raw task inputs are sampled from the same teacher model via few-shot prompting (top).
    }
    \label{fig:abstract}
\end{figure}

We apply context distillation to internalize abstract instructions and explanations. In all experiments, unless mentioned otherwise, the answer extractor is the identity function and we use few-shot prompting to sample raw task inputs (see Figure~\ref{fig:abstract}).

\paragraph{Dataset and Language Models.}

We use Natural-Instructions-V2 for all the experiments in this section. Natural Instructions is an instruction-tuning dataset of 1600+ diverse language tasks, where each task is associated with a natural language task description, input-output examples, and explanations about why certain outputs are correct or incorrect \citep{wang2022benchmarking}. 
We trained our teacher language model (TK-Instruct) by fine-tuning LM-adapted T5-11B \citep{raffel2020exploring} on the Natural Instructions training set. The training details can be found in Appendix \ref{appsec:t5_train}.
For evaluation, we select 5 tasks from the evaluation split where the teacher most significantly improves when prompted with natural language explanations, and 5 where the improvement is small.
We use Natural Instruction's official metric, Rouge-L, to calculate the performance averaged across the 10 tasks we selected.

\noindent\textbf{Hypothesis 1: context distillation can internalize abstract task instructions.}
To test this hypothesis, we defined the student template as the identity mapping, i.e., the student only sees the raw task input, while the teacher template contains the task instruction, which consists of a task description, 2 positive in-context examples, and 2-negative in-context examples (Figure \ref{fig:abstract} teacher's template).
The teacher's performance is 43.4 Rouge-L, establishing an upper bound for the student.
Before context distillation, the student's performance is 9.0, since it does not know what task it should perform. 
After context distillation, the student's performance significantly increases to 34.7. 
Context distillation successfully internalized our abstract task instructions. 
Finally, we used 11.1 times fewer inference time tokens when evaluating the student verses the teacher.

\noindent\textbf{Hypothesis 2: context distillation can learn from natural language explanations when they benefit the teacher.} 
In this experiment, we use the same teacher template as the experiment above; in contrast, in this experiment, we define a student template that is exactly the same as the teacher, but without explanations (Figure \ref{fig:abstract} student template B). 
We run context distillation to internalize the effect of natural language explanations, using the task's training split as $\mathcal{D}$ (see Appendix~\ref{appsec:distill_lang}).

Ideally, we want the student to fully internalize the effect of natural language explanations and match the teacher's performance.
To measure how much the student internalizes, we define two quantities for each task: 1) the in-context margin, which measures the performance increase after we add the explanations to the context, and 2) the distillation margin, which measures the performance increase after we perform context distillation.
We plot the margins for each task in Figure ~\ref{fig:posnegexpl}, and we observe a positive and statistically significant correlation ($r=0.75, p=1\%$).
Therefore, context distillation can learn from natural language explanations when they benefit the teacher.

Notice that for our current model, not all tasks benefit from internalizing natural language explanations.
However, we anticipate that future language models be more responsive to natural language explanations \citep{kaplan2020scaling, scheurer2022learning}, hence improving the performance of context distillation.

\noindent\textbf{Hypothesis 3: sequential distillation can overwrite past updates.} 
We study sequential distillation using four classification tasks, superglue\_copa\_text\_completion, meta\_woz\_task\_classification, tweetqa\_classification and rocstories\_title\_classification. 
We define a new task id association challenge: each task is associated with an index, and when the student model sees an index, it needs to perform the corresponding task without looking at its instruction; we train the student model to do this via simultaneous context distillation, where the student sees the task index while the teacher sees the task instruction (Figure \ref{fig:abstract} student template A). 
After we use context distillation to train the student to associate each task-id with the corresponding instruction, we shuffle the task-id association, perform context distillation again with the new shuffled association, and then evaluate the model's performance on the new association, which measures how well context distillation can overwrite previous updates.

We define two metrics for this task id association challenge. 
First, we will measure the average accuracy (correct association accuracy) of the student on the four tasks when the student is prompted with the corresponding index.
Second, it is plausible that the student can learn to associate the task input distributions, rather than the index, with the corresponding instructions.
For example, suppose that id ``[0]'' corresponds to classifying sentiment and the task input distribution is movie reviews, while ``[1]'' corresponds to whether it is sports-related and the raw input distribution is news articles, then the student might cheat by always classifying sentiment whenever it sees a movie review, regardless of what id it sees. 
Therefore, we also measure the average accuracy when we prompt the model with the wrong task id, and we want this number to be low  (wrong association accuracy): for example, if the model sees ``[0]'' and a news articles, it should have low accuracy at classifying whether it corresponds to sports, because ``[0]'' corresponds to sentiment classification.

We experimented with two variants of simultaneous distillation: 1) the ``naïve'' one, where each $\mathcal{D}_{k}$ contains only the input distribution for one single task, and 2) the ``mixed'' one, where we define each $\mathcal{D}_{k}$ as a mixture of all the raw task input distributions.
As shown in Table \ref{tab:task_assoc_avg}, under the ``naïve'' input distribution, the student model can cheat by associating the task input distribution, rather than the task id, with the task it needs to perform;
on the other hand, the ``mixed'' variant of simultaneous distillation successfully over-writes the past task id association.

\begin{table}[]
    \centering
    \begin{tabular}{|c|c|c|}
    \hline
    model & correct $\uparrow$ & wrong $\downarrow$ \\
    \hline
    Teacher & 81 & - \\
    Pre-distill Student & 49 & 48 \\
    \hline
    ``Naïve'' Post-distill Student & 68 & 61 \\
    ``Mixed'' Post-distill Student & \textbf{70} & \textbf{16} \\
    \hline
    \end{tabular}
    \caption{We apply context distillation to train the student to associate task instructions with task ids, and then apply context distillation again to overwrite the previous association. We report the correct association accuracy (``correct'') and the wrong association accuracy score (``wrong'') over all four tasks we associate. We find that the ``Mixed'' variant successfully overwrites the previous updates, as it achieves a low score of wrong index association.}
    \label{tab:task_assoc_avg}
\end{table}

\subsection{Learning from Concrete Examples} \label{sec:learning-from-examples}

\begin{figure}[t]
    \centering
    \includegraphics[width=\textwidth]{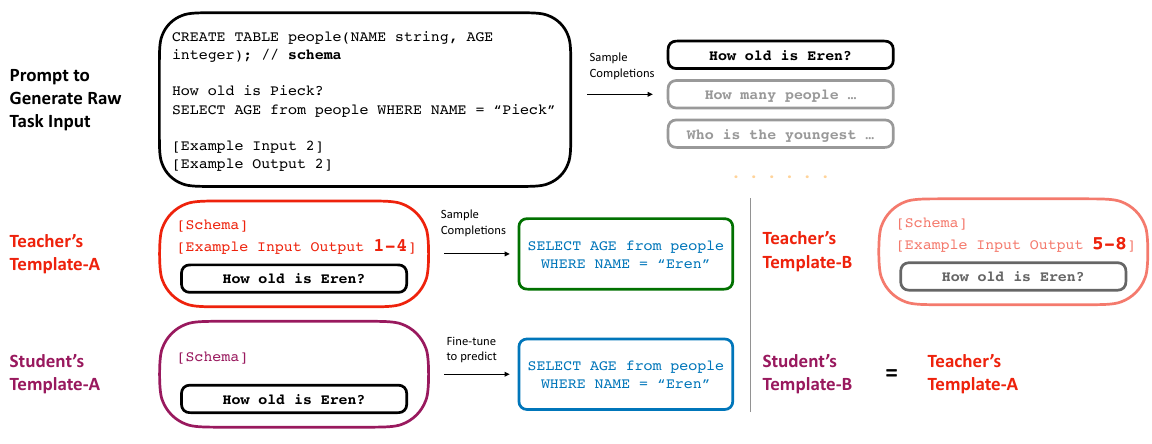}
    \caption{
    The teacher template A contains additional training examples 1-4 compared to the student template A, and context-distilling them outperforms directly learning from them with gradient descent.
    Additionally, by simultaneously applying the teacher's templates A and B (Section \ref{sec:multiple-updates}), more in-context examples (5-8) can be distilled, even though their total length might exceed the context window size.
    The raw task inputs are sampled from the same teacher model via few-shot prompting (top).
    }
    \label{fig:concrete}
\end{figure}

We show that context distillation can be used to internalize training examples, and therefore can be a potential alternative to directly learning from these examples with gradient descent;
additionally, simultaneous distillation allows the model to learn from more in-context examples when their total length exceeds the context window length.
In all experiments, the answer extractor is the identity function and we use few-shot prompting to sample raw task inputs (Figure \ref{fig:concrete}).

\paragraph{Dataset and Language Models.} 
We use the SPIDER text-to-SQL dataset ~\citep{yu-etal-2018-spider} for our experiments.
Each database in SPIDER is associated with a schema and a list of English questions annotated with SQL query.
The task is to predict the SQL query given the schema and the question. 
We choose this task because prior work~\citep{rajkumar2022evaluating} has demonstrated that SOTA models are able to effectively utilize in-context examples for this task, and we conjecture that our approach is more likely to succeed since this task requires complex cognitive reasoning. 
For the teacher language model, we chose Incoder-6.7B~\citep{fried2022incoder}, which is pre-trained on code. 
For each experiment, we randomly sampled eight text-to-SQL pairs for each database as in-context examples and evaluate on the rest by calculating the exact set match accuracy.

\paragraph{Hypothesis 4: context distillation can outperform directly learning with gradient descent.}

To test this hypothesis, we define the student template to be a database schema followed directly by the question, whereas the teacher template contains the database schema followed by four in-context examples (see Figure~\ref{fig:concrete} teacher's template-A). 
As we see in Table~\ref{tab:distillvgraddescent}, context distillation outperforms learning via gradient descent on four examples by 8.6\% in exact-set match accuracy, a margin which further increases by 0.4\% when training on eight examples. 
Therefore, context distillation can be used as an alternative to directly learning from training examples with gradient descent.

\begin{table}[]
    \centering
    \begin{tabular}{|c|c|c|}
        \hline
        Model & 4 Examples & 8 Examples \\
        \hline
        Teacher & 27.7 & 28.2 \\
        Pre-distill Student & \phantom{0}0.3 & \phantom{0}0.3 \\
        \hline
        Post-distill Student & \textbf{22.1} & \textbf{27.9} \\
        Direct Gradient Descent & 13.4 & 18.9 \\
        \hline
    \end{tabular}
    \caption{Comparing context distillation to gradient descent on the SPIDER text-to-SQL validation set. We see that context distillation outperforms directly learning via gradient descent on four examples by 8.6\% (exact set match); this margin further increases when learning from eight examples.}
    \label{tab:distillvgraddescent}
\end{table}

\paragraph{Hypothesis 5: context distillation enables learning from more training examples when their total length exceeds the context window size.}

For this experiment, we select four databases from the SPIDER training set which have particularly long database schema, such that it is possible to fit four training examples into Incoder's context window but not eight.
Therefore, we include 4 training examples in the student template (Figure \ref{fig:concrete}, student's template B.), which would lead to the best in-context learning performance ex-ante given the context window size. 
To internalize more training examples, we perform simultaneous distillation and sample teacher templates by including a random subset of four in-context examples out of eight (Figure \ref{fig:concrete}, teacher's template A and B).
After context distillation, our student achieves an exact set match of $16.2\pm0.6$, which improves over the pre-distillation student performance of $14.22\pm0.8$, hence confirming our hypothesis.

\subsection{Learning from Step-By-Step Reasoning} \label{sec:step-by-step}

We show that context distillation can effectively internalize a skills acquired by generating step-by-step reasoning, and such a skill can benefit the model for other downstream tasks. 
Unless otherwise mentioned, in this section we use an $f$ that will extract the final answer from the teacher's full output, as shown in Figure~\ref{fig:step-by-step}.

\paragraph{Dataset.} 
We define the input distribution $\mathcal{D}$ to be uniform over all possible 1 through 8 digit addition questions, identical to those used by~\citet{https://doi.org/10.48550/arxiv.2112.00114,https://doi.org/10.48550/arxiv.2203.14465}. 
We report addition accuracy for each experiment.

\paragraph{Hypothesis 6: context distillation can internalize step-by-step reasoning.}

To test this hypothesis, we obtain a teacher model that can use scratch-pad to perform step-by-step reasoning by fine-tuning the LM-adapted T5-small on a dataset of 500 addition expressions, where the model needs to first generate a scratch-pad before predicting the final answer.
We then perform context-distillation with an $f$ that extracts the final answer from the teacher's output as shown in Figure \ref{fig:step-by-step}. 
As shown in Table \ref{tab:additionscratch}, after distillation, the ability of the student to perform \textbf{direct} additions (without using scratch-pad) improves from 0\% to 94.7\%, implying that context distillation internalizes step-by-step reasoning. 

We compare this to several other transfer learning and multi-task learning baselines that use the same amount of training data in Table \ref{tab:additionscratch}. 
Under transfer learning, we first fine-tune the student model to predict the scratch pad, and then fine-tune it to directly predict the final answer.
Under multi-task learning, we fine-tune the model to predict scratch-pad and the final answer independently. 
Both variants perform significantly worse ($>20\%$) than context distillation. In addition to the gains in reasoning ability over baselines, we also saved inference time compute in our evaluations: specifically we used 8.0 times fewer tokens at inference time when evaluating the student compared to the teacher.

\begin{table}[]
    \centering
    \begin{tabular}{|c|c|c|c|c|c|c|}
        \hline
         & Teach & Pre-Dist & Post-Dist & Sc$\rightarrow$Dir & Sc+Dir \\
        \hline
        8 Digit Addition Accuracy \% & 93 & \phantom{0}0 & \textbf{95} & 72 & 61 \\
        \hline
    \end{tabular}
    \caption{Distilling addition scratchpads on T5-small. ``Teach'' refers to the teacher LM's performance using scratch-pad. ``Pre-Dist'' refers to the student's performance before distillation; ``Post-Dist'' refers to the student's performance of direct addition (without scratch-pad) after distillation;``Sc$\rightarrow$Dir''/``Sc+Dir''  
    refers to our transfer/multi-task learning baseline. Context Distillation performs the best for direct addition.}
    \label{tab:additionscratch}
\end{table}

\paragraph{Hypothesis 7: the reasoning abilities internalized by scratchpad distillation can transfer to other related reasoning tasks.}

To test this hypothesis, we distill the addition scratch-pads on our TK-Instruct model, and evaluate capability transfer to other tasks.
We obtain the teacher language model by fine-tuning TK-Instruct on a distribution of both natural instructions data and 500 addition scratchpad examples (see Appendix~\ref{appsec:stepbystep}), and we initialize the student with the original TK-Instruct checkpoint. 
For the context distillation training, we define the student template to be a description of the addition task followed by two direct answer in-context examples, and similarly the teacher template contains the task description and two in-context examples with scratch-pad answer. 
To prevent the student from catastrophically forgetting its in-context learning ability during distillation, we mix our 10k distillation data-points with a distribution of 65536 randomly selected examples from Natural Instruction-V2.

The student's accuracy on directly answering 8-digit addition questions improves from 1\% to 17\%, while the student's performance on Natural Instructions remains roughly the same (from RougeL 57 before distillation to 58 after), implying that the student did not lose its original capabilities to follow instructions.
Additionally, we evaluate the student's performance on a set of related reasoning tasks. 
We then use a template to synthesize simple questions that require knowledge to add two numbers, for example, ``A has 7 turkies. B has 2 turkies. How many turkies do they have altogether?''. 
On this synthetic dataset, the student's performance significantly increases from 17\% to 30\% after context distillation, implying that the capability to perform direct addition can transfer to other related applications.

\section{Related Work} \label{sec:related}

\paragraph{Prompting and Instruction Tuning.} 
Many recent works show that language models can learn from abstract task definitions \citep{zhong-etal-2021-adapting-language, mishra-etal-2022-cross, wei2022finetuned, sanh2022multitask}, natural language explanations \citep{scheurer2022learning, wang2022benchmarking}, and concrete in-context examples \citep{min-etal-2022-metaicl, chen-etal-2022-meta}. 
We anticipate the in-context learning performance to improve further in the future \citep{kaplan2020scaling}, thus increasing the upper bound of what context distillation can achieve.

\paragraph{Scratch Pad.}
Many recent works show that language models perform better when it is required to generate a chain of reasoning steps before outputting the final answer \citep{zhou2021think, nye2021show, cobbe2021training, wei2022chain, zhou2022least, lewkowycz2022solving}.
We anticipate context distillation to be increasingly useful, as the research community starts to tackle more difficult problems, which require more sophisticated skills and have longer problem descriptions and reasoning chains. 

\paragraph{Distillation.}
There has been a large literature on distilling knowledge in a neural network \citep{hinton2015distilling, adriana2015fitnets, liu2019improving, yang2020model, 9156610}.
Most related to our work, different sub-variants of context distillation have been independently discovered by different researchers.
\citet{wang2021towards} emphasizes the aspect of creating a dataset without any human annotations and uses a language model to generate task inputs and their labels. 
\citet{https://doi.org/10.48550/arxiv.2206.11349,https://doi.org/10.48550/arxiv.2112.00861} formulated the method of context distillation (also referred to as prompt injection), which distills a fixed input prompt; their method is a special case of our framework with an identity student template and an identity output selector.
Additionally, they focused more on the benefit of saving computational resources, while we considered it as a general learning method. 
Concurrent to our our work, \citet{anonymous2023large} focuses on internalizing step-by-step reasoning, and they corroborated our findings on much large models and a much wider range of datasets.

\section{Conclusion}
We present context distillation as a general method for learning, which can internalize abstract statements, concrete examples, and step-by-step reasoning. 
Given that 1) it is general and delivers strong performance, 2) future models will have stronger in-context learning capability, and 3) future tasks will have longer descriptions and reasoning chains, we anticipate our methods to be increasingly useful in the future.

\section{Acknowledgement}
We thank Jacob Steinhardt, Sergey Levine, Kevin Yang, Nicholas Tomlin, and other members of the Berkeley NLP group for their helpful feedback.
We thank the TPU Research Cloud (TRC) program for providing computational resources.

\bibliography{iclr2023_conference}
\bibliographystyle{iclr2023_conference}
\appendix
\pagebreak

\section{Other applications} \label{sec:other-applications}

We show that context distillation could be applied to other applications, such as controlled text generation or factual knowledge editing. 

\subsection{Controlled Text Generation via Context Distillation}

By distilling prompts that describe the desired behavior (e.g. don't output toxic language or only generate positive sentiment text), we can exert a form of control in language models. 
To test this, we prompt our TK-Instruct model to complete negative movie reviews from the IMDB sentiment classification dataset~\citep{maas-etal-2011-learning}.

The teacher template contains the instruction to complete the movie review specifically to end the review on a positive note, and the student template contains the instruction just to complete the review without any other specification. 
After distillation, the student should learn to internalize the abstract control preference for positive movie reviews. 

We evaluate our models by asking GPT-3 to verify that our model's generation is indeed positive.
As shown in Table~\ref{tab:imdb_control}, context distillation allows us to control our language model to generate positive reviews.

\begin{table}[]
    \centering
    \begin{tabular}{|c|c|c|c|c|c|c|c|c|c|}
        \hline
          & \multicolumn{3}{|c|}{teacher} & \multicolumn{3}{|c|}{Pre-distillation Student} & \multicolumn{3}{|c|}{Post-Distillation Student} \\
         \hline
         & sent & RougeL & ent & sent & RougeL & ent & sent & RougeL & ent \\
        \hline
        positive & \textbf{94.0} & \textbf{4.8} & \textbf{13.0} & 0.24 & 0.6 & 2.5 & \textbf{0.96} & 4.8 & \textbf{14.4} \\
        \hline
        neutral & 0.54 & 3.6 & 9.9 & 0.24 & 0.6 & 2.5 & 0.34 & \textbf{7.0} & 14.2 \\
        \hline
    \end{tabular}
    \caption{Controlled generation via-context distillation. The ``sent'' column reports the average sentiment score of generations from the model, and the ``ent'' column refers to the model's estimated output entropy in nats. We see that by distilling the positive control instruction into the model, we can obtain a model which outputs greater positive sentiment text without significantly sacrificing output coherence (RougeL) or output diversity (entropy).}
    \label{tab:imdb_control}
\end{table}

\subsection{Factual Knowledge Editing with Context Distillation}

Context distillation also provides a natural way to edit the factual knowledge implicitly internalized by language models, by distilling a prompt that states a new or edited declarative fact. 
This is in contrast to prior works~\citep{https://doi.org/10.48550/arxiv.2110.11309,https://doi.org/10.48550/arxiv.2104.08164,https://doi.org/10.48550/arxiv.2202.05262} on fact editing, which instead perform a constrained optimization procedure that directly learns an edit to the model parameters, corresponding to the factual knowledge update.

We use the challenging Counterfact dataset from ~\citet{https://doi.org/10.48550/arxiv.2202.05262} to test our method's fact editing ability. 
Each instance of the Counterfact task involves editing the object of a given factual relation. 
Ideally, the language models should consistently apply the new fact under significant paraphrases to the original relation and context changes; the model should also not update the unrelated knowledge. 
To measure this, the Counterfact dataset provides a set of paraphrase prompts (significant paraphrases of the original fact, which the LM should consistently edit), neighborhood prompts (un-related facts that share the same object as the original pre-edit fact, on which the model should not change their predictions), and attribute prompts (un-related facts which share the same object ad the new post-edit fact, which should not change).

We perform fact editing experiments on our TK-Instruct model. For a randomly selected fact corresponding to each of the 34 unique relations in the Counterfact dataset, we synthesize a teacher template, which includes a description of the fact to be edited and instructions not to edit unrelated facts. 
We also use GPT-3 to help us generate a new attribute, paraphrase, and neighborhood prompt for each fact edit to use as demonstrations of desired behavior in the prompt, alongside a natural language explanation of why the fact edit is or is not applied in each case. 
We generate the inputs $P(x)$ using a few-shot prompt to TK-Instruct.

In Table~\ref{tab:fact_edit}, we evaluate our model on the set of paraphrase and neighborhood prompts in the dataset. We report both the average score – $\mathbbm{1}[P(\text{correct object}) > P(\text{incorrect object})]$ – and the average magnitude – $P(\text{correct object}) - P(\text{incorrect object})$ – under the language model, where $P(\text{correct object})$ and $P(\text{incorrect object})$ are the probability of the correct and incorrect objects under the model, when conditioned on the relevant subject and relation. 
We see that context distillation is largely able to recover the fact editing performance of the teacher and performs comparably in absolute score to current SOTA approaches to fact editing – ROME~\citep{https://doi.org/10.48550/arxiv.2202.05262} and MEND~\citep{https://doi.org/10.48550/arxiv.2110.11309}.
Notice that we show these numbers for the readers to interpret our result and we emphasize that our method is not directly comparable to theirs, since we use much more computational resources.  

Unfortunately, our prompted TK-Instruct teacher model performs poorly on neighborhood prompts, which also leads to poor performance of the student model. 
We expect this issue to be largely resolved by context distilling larger and more capable language models. 
To demonstrate this, we evaluate GPT-3 on this task with the same prompt, which we can see in Table~\ref{tab:fact_edit} performs much better on these neighborhood prompts.
While we cannot perform context distillation on GPT-3 due to limitations in OpenAI's API, we expect these improvements to carry over to the distilled model.

\begin{table}[]
    \centering
    \begin{tabular}{|c|c|c|c|c|c|c|}
        \hline
        method & \multicolumn{2}{|c|}{\textbf{paraphrase}} &  \multicolumn{2}{|c|}{\textbf{neighborhood}} \\
        \hline
        & score & magnitude & score & magnitude \\
        \hline
        Teacher & 73 & 29 & 58 & 8 \\
        Pre-distill student & 34 & -3 & 80 & 4 \\
        Post-distill student & 79 & 28 & 48 & -2 \\
        \hline
        GPT-3 & 65 & \phantom{0}3 & 75 & 17 \\
        MEND & 65 & 12 & 38 & -12 \\
        ROME & 89 & 33 & 74 & \phantom{0}4 \\
        \hline
    \end{tabular}
    \caption{Performance of context distillation on fact editing. We see that context distillation is able to recover the paraphrase score of the teacher but slightly under-performs in neighborhood score.}
    \label{tab:fact_edit}
\end{table}
\section{Experiment Details}
\subsection{Instruction-tuned T5-11b.} \label{appsec:t5_train}

Following the procedure of~\citep{https://doi.org/10.48550/arxiv.2204.07705}, we fine-tuned the 11B T5 LM-adapted model~\citep{https://doi.org/10.48550/arxiv.1910.10683}, on Natural Instructions V2~\citep{https://doi.org/10.48550/arxiv.2204.07705},
a large dataset of 1600+ language tasks which includes, for each task, a task description, positive and negative in-context examples, and natural language explanations of why the output is right or wrong for each of the in-context examples. 
Prior work~\citep{https://doi.org/10.48550/arxiv.2204.07705} has used this dataset to train instruction-tuned models on prompts consisting of only 2-positive examples or only 2-positive and negative examples with explanations. 
To maximize the flexibility of our instruction-tuned model, we instead instruction-tuned on a distribution of randomized prompts, which consist of randomly chosen 0 to 3 positive examples, 0 to 3 negative examples, and whether there is an explanation or not. We trained the model for 9728 steps with a batch size of 16 and AdamW optimizer on 32 TPU-V3 cores. 
The model achieves a RougeL score of $58$ on the 2 positive, 2 negative with explanation test split, of unseen natural instructions tasks.

\subsection{Finetuning Language Model Implementation Details}

We run all experiments on 32 TPU-V3 cores, with the model parameters and optimizer states for fine-tuning sharded equally across all cores. Our codebase is built in Jax~\citep{jax2018github,50530} using the PJIT function to handle the model parallel training and inference.

\subsection{General Experiment Details} \label{appsec:general_exp}

For all experiments, except where denoted otherwise, we distill on 4096 examples for 1 epoch with batch size 16 with AdamW optimizer. We use a learning rate of 1e-5 with TK-Instruct and 1e-4 with Incoder. We use 0 weight decay for all experiments.

\subsection{Distilling Abstract Statements Details} \label{appsec:distill_lang}

\begin{figure}
    \centering
    \includegraphics[scale=0.5]{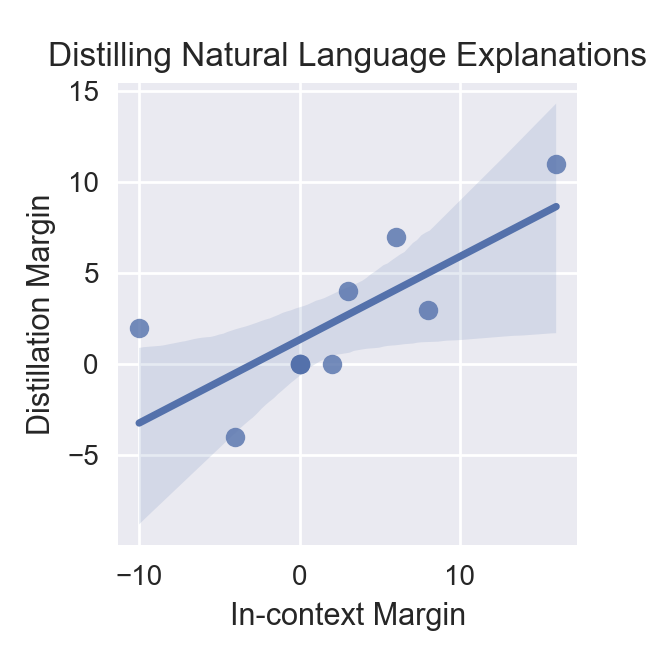}
    \caption{Distilling explanations for 10 tasks from the Natural Instructions V2 test set. Each point corresponds to a task in the figure. ``In-context Margin'' quantifies how much the explanations help the teacher by measuring the difference in RougeL of TK-Instruct with and without explanations on each task, and ``Distillation Margin'' quantifies how much the student learns from the teacher by measuring the difference between the RougeL score of the student after distillation and the RougeL score of TK-Instruct without explanations on each task. We observe a positive correlation between the utility of the explanations to the teacher and how much our student learns.}
    \label{fig:posnegexpl}
\end{figure}

\paragraph{Learning from natural language explanations details (hypothesis 2).} We present a scatter plot of the ``In-context Margin'' verses the ``Distillation Margin'' for our experiment on learning from natural language explanations in Figure~\ref{fig:posnegexpl}.

\paragraph{Distilling task id associations details (hypothesis 3).} We present detailed, per-task results for our task id association experiment in Table~\ref{tab:task_assoc1}.

\begin{table}[]
    \centering
    \begin{tabular}{|c|c|c|c|c|c|c|c|c|}
        \hline
        model & \multicolumn{2}{|c|}{SG COPA} & \multicolumn{2}{|c|}{Meta-Woz} & \multicolumn{2}{|c|}{Tweet-QA} & \multicolumn{2}{|c|}{ROCStories} \\
        & c $\uparrow$ & w $\downarrow$ & c $\uparrow$ & w $\downarrow$ & c $\uparrow$ & w $\downarrow$ & c $\uparrow$ & w $\downarrow$ \\
        \hline
        Teacher & 74 & - & 74 & - & 80 & - & 97 & - \\
        Pre-distill Student & 63 & 62 & \phantom{0}0 & \phantom{0}0 & 51 & 51 & 80 & 78 \\
        \hline
        ``Naïve'' Post-distill Student 1 & 70 & 78 & 26 & 13 & 59 & 47 & \textbf{98} & 99 \\
        ``Mixed'' Post-distill Student 1 & \textbf{79} & \textbf{15} & \textbf{38} & \phantom{0}0\textbf{0} & \textbf{63} & \phantom{0}\textbf{0} & 96 & \textbf{46} \\

        \hline
        ``Naïve'' Post-distill Student 2 & \textbf{77} & 85 & 34 & 29 & 61 & 31 & \textbf{98} & 97 \\
        ``Mixed'' Post-distill Student 2 & 76 & \textbf{18} & \textbf{47} & \phantom{0}\textbf{0} & \textbf{63} & \phantom{0}\textbf{0} & 94 & \textbf{44} \\
        \hline
    \end{tabular}
    \caption{We distill the student to associate 4 task instructions with task ids, we then override this association by permuting the task ids. ``Student 1'' refers to the the student after this first step of task-id association, and ``Student 2'' refers to the student after overriding the task-id association. We see that in general ``Mixed'' students successfully distill the task-id associations. We report the correct association accuracy (``c'') and the wrong association accuracy score (``w'') over all four tasks we associate. We find that the ``Mixed'' variant successfully overwrites the previous updates, as it achieves a low score of wrong index association.}
    \label{tab:task_assoc1}
\end{table}

\subsection{Distilling Concrete Examples Details} \label{appsec:distill_examples}

\paragraph{Gradient descent details (hypothesis 4).} For gradient descent we fit all training examples into a single batch and train for a 25 epochs using the AdamW optimizer with a learning rate of 1e-5. We report the performance of the epoch with the highest average exact set match score across all databases.

\paragraph{Distilling long contexts details (hypothesis 5).}

To estimate the per-token log-probabilities of the $y$ sampled from the teacher ensemble for distillation, we average the probabilities of each $y$ under 8 teacher prompts. We estimate the teacher performance by performing greedy decoding on the ensemble of 8 prompts uniformly sampled from the set of all 4 choose 8 teacher prompts. For each database, we distill two students with different in-context examples in the prompt, and we report the average exact-set match accuracy for both of these students.

\subsection{Distilling Step-by-Step Reasoning Details} \label{appsec:stepbystep}

\paragraph{Distilling scratchpads with T5-small (hypothesis 6).} All baselines were trained for 1000 epochs – except ``Scratchpad then Direct'' which was trained for 1000 epochs to predict scratchpads and then 1000 epochs to directly predict the answer – with a batch size of 8, a learning rate of 3e-4, and AdamW optimizer. We report performance at the end of training on 10k unseen addition problems.

\paragraph{Transfering step-by-step reasoning details (hypothesis 7).} Since TK-Instruct cannot successfully do scratchpad addition from a few shot prompt, to initialize the teacher, we first fine-tune TK-Instruct on the same distribution of 500 scratchpad examples from the previous experiment mixed in with 4096 randomly selected examples from the ``2 positive'' split of Natural Instructions-V2 training set, such that the teacher doesn't lose its ability to respond to prompts. We train the teacher for 32 epochs, at which point it achieves 97\% accuracy on 2000 held-out scratchpad addition problems. Our student is trained for 3 epochs on a dataset consisting of 10k distillation examples mixed in with 65536 randomly selected examples from the same split of Natural-Instructions-V2 training set as the teacher.

\end{document}